\documentclass{article}

    \PassOptionsToPackage{numbers, compress}{natbib}
 \usepackage[preprint]{neurips_2025}

\usepackage[utf8]{inputenc} 
\usepackage[T1]{fontenc}    
\usepackage{hyperref}       
\usepackage{url}            
\usepackage{booktabs}       
\usepackage{amsfonts}       
\usepackage{nicefrac}       
\usepackage{microtype}      
\usepackage{xcolor}         

\usepackage{amsmath}
\usepackage{amsthm}
\newtheorem{theorem}{Theorem}

\newtheorem{proposition}{Proposition}
\newtheorem{lemma}{Lemma}
\newtheorem{definition}{Definition}
\usepackage{color}
\usepackage{colortbl}
\usepackage{wrapfig}
\usepackage{subcaption}

\usepackage{booktabs}						
\usepackage{multirow}						
\usepackage{amsfonts}						
\usepackage{graphicx}						
\usepackage{duckuments}						

\newcommand\ie{\textit{i.e.,}}
\newcommand\eg{\textit{e.g.,}}

\newcommand{\beq}{\begin{equation}}
\newcommand{\eeq}{\end{equation}}
\newcommand{\beqnn}{\begin{equation*}}
\newcommand{\eeqnn}{\end{equation*}}
\newcommand{\beqy}{\begin{eqnarray}}
\newcommand{\eeqy}{\end{eqnarray}}
\newcommand{\beqynn}{\begin{eqnarray*}}
\newcommand{\eeqynn}{\end{eqnarray*}}
\newcommand{\bit}{\begin{itemize}}
\newcommand{\eit}{\end{itemize}}
\newcommand{\ben}{\begin{enumerate}}
\newcommand{\een}{\end{enumerate}}
\newcommand{\bex}{\begin{example}}
\newcommand{\eex}{\end{example}}

\newcommand{\balg}[1]{\begin{algorithm} \caption{#1}}
\newcommand{\ealg}{\end{algorithm}}

\newcommand{\balgc}{\begin{algorithmic}[1]}
\newcommand{\ealgc}{\end{algorithmic}}

\newcommand{\bary}{\begin{array}}
\newcommand{\eary}{\end{array}}
\newcommand{\bmx}{\begin{bmatrix}}
\newcommand{\emx}{\end{bmatrix}}
\newcommand{\bsmx}{\left[\begin{smallmatrix}}
\newcommand{\esmx}{\end{smallmatrix}\right]}
\newcommand{\bmxc}[1]{\left[\begin{array}{@{}#1@{}}}
\newcommand{\emxc}{\end{array}\right]}
\newcommand{\bcn}{\begin{center}}
\newcommand{\ecn}{\end{center}}

\title{Exploring Heterophily in Graph-level Tasks}
\workshoptitle{New Perspectives in Graph Machine Learning}

\newif\ifuniqueAffiliation

\ifuniqueAffiliation 
\else
\usepackage{authblk}

\author{
Qinhan Hou$^{1,2}$, Yilun Zheng$^{3}$, 
Xichun Zhang$^{2}$, Sitao Luan$^{4,5,*}$, Jing Tang$^{2,*}$\\
\small $^1$Doctoral Program of Computer Science, University of Helsinki, Helsinki, Finland \\
\small $^2$Research Program in Systems Oncology, Faculty of Medicine, University of Helsinki, Helsinki, Finland\\
\small $^3$Centre for Information Sciences and Systems, Nanyang Technological University, Singapore\\
\small $^4$University of Montreal, Canada\\
\small $^5$Mila, Quebec Artificial Intelligence Institute, Canada\\
\small $^*$Corresponding author, main supervision
}
\fi

\begin{document}
\maketitle

\begin{abstract}
While heterophily has been widely studied in node-level tasks, its impact on graph-level tasks remains unclear. We present the first analysis of heterophily in graph-level learning, combining theoretical insights with empirical validation.
We first introduce a taxonomy of graph-level labeling schemes, and focus on motif-based tasks within local structure labeling, which is a popular labeling scheme. Using energy-based gradient flow analysis, we reveal a key insight: unlike frequency-dominated regimes in node-level tasks, motif detection requires mixed-frequency dynamics to remain flexible across multiple spectral components. Our theory shows that motif objectives are inherently misaligned with global frequency dominance, demanding distinct architectural considerations. Experiments on synthetic datasets with controlled heterophily and real-world molecular property prediction support our findings, showing that frequency-adaptive model outperform frequency-dominated models. This work establishes a new theoretical understanding of heterophily in graph-level learning and offers guidance for designing effective GNN architectures.
\end{abstract}

\section{Introduction}

Graph Neural Networks (GNNs) have achieved success in learning from
graph-structured data, demonstrating strong performance across diverse domains
including social networks~\cite{kipf2016classification, luan2019break} and molecular property prediction~\cite{zhang2018graph, zhou2018graph}. Many popular GNN architectures, such as Graph Convolutional Networks (GCNs)~\cite{gilmer2017neural}, are designed under the homophily assumption, \ie{} connected nodes tend to share similar features or labels~\cite{mcpherson2001birds, hamilton2017representation, luan2023graph}. However, many real-world graphs exhibit heterophily, where neighboring nodes have dissimilar characteristics~\cite{zhu2023heterophily, luan2022revisiting}. While the challenges posed by
heterophily have been extensively studied on node-level tasks~\cite{luan2021heterophily, lu2024flexible, luan2024heterophilic, zheng2024rethinking, zheng2024missing, zheng2025let}, its impact on graph-level tasks remains poorly understood.

This work is the first to study heterophily in graph-level tasks. We introduce a taxonomy that classifies such tasks into three types by their labeling mechanisms, focusing on motif-based tasks where labels depend on discriminative subgraphs (\textit{motifs}). From an energy and gradient flow perspective~\cite{digiovanni2023understanding}, our analysis shows that graph-level tasks have distinct frequency preferences from node-level tasks, as motif detection misaligns with the global nature of low- and high-frequency dominant regimes. This misalignment challenges the effectiveness of GNN under heterophily settings. We provide both theoretical and empirical evidence, offering new insights into heterophily’s role in graph-level prediction and guiding the design of more adaptive GNNs.

\section{Task Taxonomy, Notations and Background}
\subsection{Task Taxonomy}
\label{sec:taxonomy}
Compared to node-level tasks, the labeling schemes of graph-level tasks make it challenging to establish a simple and general relation between graph labels and certain graph properties. To enable a simplified and systematic discussion, we categorize the labeling schemes into three main types based on the following criterion derived from practical applications.

\textbf{Aggregated Node Features.} In this scenario, graph labels are primarily determined by aggregated node features. For instance, a graph may be assigned to a particular class if the average value of a specific node feature across all nodes exceeds a given threshold, or if a certain proportion of nodes belong to a particular latent class, \textit{\eg{}} online community detection based on aggregated user demographics~\cite{newman2013community, fortunato2016community, chen2019supervised}.

\textbf{Local Structure.} Labels depend on local structural patterns and node-level features. For example, a label may be assigned based on the presence of a specified number of particular motifs (\textit{\eg{}} triangles) within the graph. These motifs may predominantly consist of either homophilic or heterophilic nodes, \textit{\eg{}} molecular classification based on the presence of specific pharmacophores or toxic substructures~\cite{Debnath1992structure, wu2018moleculenetbenchmarkmolecularmachine, kazius_derivation_2005, toivonen2003statistical}.

\textbf{Global Structure.} In this case, labels are determined by global structural properties of the graph, such as its diameter or overall connectivity. The label thus reflects a purely structural characteristic of the graph, \eg{} metabolic network categorization of different organisms based on global connectivity
patterns, such as scale-free vs. random network topologies~\cite{zhu_structural_2005, mazurie_evolution_2010, ramon_functional_2023}.

To give a more intuitive understanding, we list some real-world applications according to these three types of tasks in the Appendix \ref{ref:real-world-examples}. Note that the above three categories are not exclusive, and a graph can be classified by mixture criteria. To simplify the discussion, we focus on the local structure labeling in this work, which is common in real world.

\subsection{Energy-Based Framework for Understanding GNN Dynamics}
Recent work by Di Giovanni et al.~\cite{digiovanni2023understanding} 
provides a rigorous framework for analyzing GNNs
as dynamical systems that minimize a generalized energy functional.
This framework reveals that under certain conditions, the training dynamic of GNN will lead to a global frequency-dominated regime, leading to a bipolar convergence of node features. We review this framework briefly.

\paragraph{GNN Dynamics as Gradient Flow} Consider an undirected and connected graph $G=(V,E)$, where nodes
$v = \{v_1, v_2, \dots, v_n\}\in V$ have features $\{\textbf{f}_i \in\mathbb{R}^d: v_i\in V\}$, and edge set denotes $E \subset V \times V$. The feature matrix $\textbf{F} \in \mathbb{R}^{n\times d}$ consists of $\textbf{f}_i$ as its
rows. According to~\cite{bronstein2021geometricdeeplearninggrids}, Message Passing Neural Networks (MPNNs)~\cite{gilmer2017neural} update the layer $t+1$ via:
\begin{equation}
	\label{update_eq}
	\dot{\textbf{F}} = \textbf{F}(t+1) - \textbf{F}(t) = \sigma(-\textbf{F}(t) \mathbf{\Omega}_t + \textbf{A} \textbf{F}(t) \textbf{W}_t-\textbf{F}(0)\tilde{\textbf{W}}_t)
\end{equation}
where $\Omega_t$, $\textbf{W}_t$, and $\tilde{\textbf{W}}_t$ are learnable matrices performing feature transformations, $\sigma$ is the non-linear activation function, and $\textbf{A}$ is the adjacency matrix which aggregates neighbor information.

A \textbf{gradient flow} is a special dynamical system which is defined by an ordinary differential equation $\dot{\textbf{F}}(t)=-\nabla \mathcal{E}(\textbf{F}(t))$. The dynamic in Eq.\ref{update_eq} corresponds to a gradient flow of the energy functional:
\begin{equation}
	\label{general_energy}
	\mathcal{E}_{\mathbf{\theta}}(\mathbf{F}) \;=\;
	\sum_i \langle \mathbf{f}_i, \mathbf{\Omega} \mathbf{f}_i \rangle\vphantom{\sum_{i,j}}
	\;-\;
	\sum_{i,j} A_{ij} \langle \mathbf{f}_i, \mathbf{W} \mathbf{f}_j \rangle\vphantom{\sum_{i,j}}
	\;+\;
	\varphi^{0}(\mathbf{F}, \mathbf{F}(0))\vphantom{\sum_{i,j}}
\end{equation}
given the conditions that the weight matrices $\mathbf{\Omega}$ and $\mathbf{W}$ are symmetric~\cite{digiovanni2023understanding}, and $\varphi^{0}$ is a pre-defined function to calculate the distance between $\textbf{F}$ and the source $\textbf{F}(0)$. Note that the Dirichlet energy $\mathcal{E}^{Dir}(\textbf{F}(t)):=\frac{1}{2}\sum_{(i,j)\in E}\|\nabla\textbf{F}(t)\|^2$ is a special case of $\mathcal{E}_\mathbf{\Theta}$ when $\mathbf{\Omega}=\textbf{W}=\textbf{I}_d$ and ${\varphi^0}=0$.



\paragraph{Asymptotic Frequency-Dominated Regimes} This energy framework reveals that linear GNNs converge to one of two asymptotic behaviors, characterized by the relationship between the graph Laplacian spectrum and the eigenvalues of the weight matrix $\textbf{W}$. Let $\mathbf{\Delta} = \mathbf{I} - \mathbf{D}^{-1/2}\mathbf{A}\mathbf{D}^{-1/2}$ be the normalized Laplacian with ordered eigenvalues $0 = \boldsymbol{\lambda}_0 \le \dots \le \boldsymbol{\lambda}_{n-1}$ and corresponding eigenvectors $\{\boldsymbol{\phi}_0, \dots, \boldsymbol{\phi}_{n-1}\}$. We give the symmetric definition for the frequency-dominated dynamics and a theorem to decide the dynamics in a simplified version of MPNN.

\begin{definition}[Frequency-Dominated Dynamics]
The dynamics of a GNN depends on the limiting behavior of the normalized Dirichlet energy $\mathcal{E}^{Dir}(\mathbf{F}(t))/\|\mathbf{F}(t)\|^2$ 
. If it converges to the eigenvalue $\boldsymbol{\lambda}_0$, we call the dynamic Low-Frequency-Dominant (LFD). Conversely, if it converges to the eigenvalue $\boldsymbol{\lambda}_{n-1}$, we call it High-Frequency-Dominant (HFD).
\end{definition}

We refer a lemma to illustrate the condition to decide whether the MPNN is HFD or LFD.

\begin{lemma} [Theorem 4.3 in \cite{digiovanni2023understanding}]
    Given a continuous MPNN of the form $\dot{\textbf{F}}(t)=\textbf{A}\textbf{F}(t)\textbf{W}$, let $\boldsymbol{\mu}_0 \le \boldsymbol{\mu}_1 \le \dots \le \boldsymbol{\mu}_{d-1}$ be the eigenvalues of $\textbf{W}$. If $|\boldsymbol{\mu}_0|(\boldsymbol{\lambda}_{n-1}-1)>\boldsymbol{\mu}_{d-1}$, then for almost every $\textbf{F}(0)$, the MPNN is HFD. Conversely, if $|\boldsymbol{\mu}_0|(\boldsymbol{\lambda}_{n-1}-1)<\boldsymbol{\mu}_{d-1}$, then for almost every input $\textbf{F}(0)$, the MPNN is LFD.
\end{lemma}

From the lemma, the dynamic of the network depends on the biggest and lowest eigenvalues of $\textbf{W}$: $\boldsymbol{\mu}_0$ and $\boldsymbol{\mu}_{d-1}$. Since the network is a gradient flow along the energy functional described in Eq.~\ref{general_energy}, the network is trained to minimize the energy. We will discuss below how will this energy decreasing affect $\textbf{W}$ and its eigenvalues, especially under heterophily situation.


\subsection{Graph Heterophily in Energy-based Framework}
\label{graph_heterophily_in_energy_framework}
\textbf{Heterophily} refers to the tendency of connected nodes to have dissimilar features or labels, in contrast to homophily where neighboring nodes are similar~\cite{zhu2023heterophily, luan2024heterophilic}. Within the energy functional framework, heterophily has a direct correspondence to the spectral behavior of GNNs.

Recall the weight matrix $\textbf{W}$ in Eq.~\ref{general_energy}, it can be rewritten as $\textbf{W}=\mathbf{\Theta}_{+}^{\top}\mathbf{\Theta}_{+}-\mathbf{\Theta}_{-}^{\top}\mathbf{\Theta}_{-}$ by decomposing it into components with positive and negative eigenvalues (see the derivation at~\ref{eigen_decomposition})
. The pairwise interaction term $\sum_{i,j} A_{ij} \langle \mathbf{f}_i, \mathbf{W} \mathbf{f}_j \rangle$ in Eq.~\ref{general_energy} captures the relationship between connected nodes. When $\textbf{W}$ has predominant positive eigenvalues, which leads to $[\mathbf{\Theta}_{+}^{\top}\mathbf{\Theta}_{+}]_{i,j} \gg [\mathbf{\Theta}_{-}^{\top}\mathbf{\Theta}_{-}]_{i,j}$, the maximization of the pairwise interaction term will encourage neighboring nodes to have aligned representations. This naturally smooths the features across connected nodes, promotes GNN to LFD dynamics, so that benefits GNN performance on homophilic graphs.

Conversely, when $\textbf{W}$ has significant negative eigenvalues (\ie{} $\mathbf{\Theta}_{-}^{\top}\mathbf{\Theta}_{-}$ dominates), the optimization of the pairwise interaction term will encourage connected nodes to have \textit{anti-aligned} or dissimilar representations. This will drive the system toward HFD dynamics, where high-frequency components dominate, and sharp differences emerge between connected nodes. And the GNN will end up with an energy landscape that favors heterophilic patterns.

The eigenvalue decomposition $\textbf{W}=\mathbf{\Theta}_{+}^{\top}\mathbf{\Theta}_{+}-\mathbf{\Theta}_{-}^{\top}\mathbf{\Theta}_{-}$ thus reveals a fundamental trade-off: the energy framework naturally biases GNNs toward either global homophily (LFD) or global heterophily (HFD), but not both simultaneously 
. This global preference 
creates challenges for tasks that require \textit{local adaptation}—where some regions of the graph exhibit homophilic patterns (\eg{} within motif instances) while others exhibit heterophilic patterns (\eg{} at motif boundaries) 
.


\section{Motif Detection Requires Mixed-Frequency Dynamics}
Given that the energy framework naturally leads GNNs to frequency-dominated regimes, in this section, we will show that graph-level motif detection represents a fundamentally different class of problems which cannot be solved by either pure low-frequency or high-frequency dynamics. 

\subsection{Shift Motif Detection to Node-level Task}
A given motif $M = (V_M, E_M)$ is a connected graph pattern, where $V_M \subseteq V$ and $E_M = \{(u,v) \in E : u, v \in V_M\}$ 
. There exists a graph isomorphism $\psi: V_M \to V'$ such that $(u,v) \in E_M$ if and only if $(\psi(u), \psi(v)) \in E'$. While often framed as a graph-level problem (\ie{} determining if a graph contains a motif), its inherent dependency on local structure makes it easy to be formulated as a node-level problem. This perspective can shift the objective from a graph-level task to a node-level task, which is to identify all the nodes that belong to the motif. We define the objective of the motif detection task as a node-level task:

\begin{definition}[Node-Level Motif Detection]
\label{node-level-motif-detection}
For a graph $G = (V, E)$ and motif $M$, we assign binary labels $y_i \in \{0,1\}$ to each node $i \in V$, where $y_i = 1$ if the node $i$ is part of any subgraph isomorphic to $M$, and $0$ otherwise. The task is to learn $f_{node}: G \rightarrow \{0,1\}^{|V|}$ that predicts $\mathbf{y} = (y_1, \ldots, y_{|V|})$ using a GNN encoder $E_{node}$ and classifier.
\end{definition}

Furthermore, we prove the equivalence of the task objectives in different levels.

\begin{proposition} [Equivalence of Node-Level and Graph-Level Motif Detection.]
\label{equivalence_between_tasks}
For graph $G = (V, E)$ and motif $M$, the following objectives are equivalent: (i) detect if $G$ contains a subgraph isomorphic to $M$; (ii) detect whether $\exists i \in V$ such that $y_i = 1$ in the node-level task.
\end{proposition}

\paragraph{Heterophily Patterns in Motif Detection} 
\label{heteorphily_patterns_in_motif_detection}
The node-level motif detection task reveals a fundamental heterophily challenge that distinguishes graph-level tasks from node-level tasks. Unlike traditional node classification where heterophily is characterized globally across the entire graph, motif detection requires handling \textit{spatially-varying} heterophily patterns.

Effective motif detection demands three distinct connectivity behaviors. First, \textbf{intra-motif homophily} requires nodes within the same motif instance to have similar representations for consistent detection ($y_i = y_j = 1$ for $i,j \in V_M$). Second, \textbf{inter-motif heterophily} necessitates strong representational boundaries between motif participants and non-participants ($\mathbf{f}_i \neq \mathbf{f}_j$ for $i \in V_M, j \notin V_M$ when $(i,j) \in E$). Finally, \textbf{context-dependent adaptation} means the same edge $(i,j)$ may require homophilic smoothing if both nodes are motif participants, or heterophilic sharpening if they represent a motif-background boundary.

This spatial heterogeneity in connectivity requirements creates a fundamental mismatch with the energy framework's global frequency preferences, as we demonstrate below.


\subsection{Theoretical Incompatibility with Frequency-Dominated Regimes}

\paragraph{From Heterophily to Mixed-Frequency Requirements}
The spatially-varying heterophily patterns required for motif detection directly translate to mixed-frequency requirements in the spectral domain. Recall from Section~\ref{graph_heterophily_in_energy_framework} that LFD dynamics correspond to global homophily (feature smoothing), while HFD dynamics correspond to global heterophily (feature sharpening). However, motif detection requires \textit{both} behaviors simultaneously but at different spatial locations.

Specifically, the optimal node representation $\mathbf{f}^*_i$ for motif detection must satisfy conflicting spectral requirements: (\textbf{i}) low-frequency components are needed within motif instances to maintain intra-motif consistency ($\mathbf{f}^*_i \approx \mathbf{f}^*_j$ for $i,j \in V_M$), (\textbf{ii}) high-frequency components are essential at motif boundaries to create discriminative separation ($\|\mathbf{f}^*_i - \mathbf{f}^*_j\|$ large for $i \in V_M, j \notin V_M$), and (\textbf{iii}) medium-frequency components may be required for motifs of specific structural scales.

The energy functional's global optimization toward either LFD or HFD regimes cannot accommodate this spatial heterogeneity. A purely LFD approach would blur motif boundaries through over-smoothing, while a purely HFD approach would fragment intra-motif coherence through excessive sharpening. This fundamental incompatibility explains why frequency-dominated GNNs struggle with motif detection across different heterophily settings.

To give a theoretical analysis on the incompatibility with the frequency-dominated regimes, we first draw a lemma that the performance of GNN on motif-based graph-level tasks is upper bounded by its performance on the node-level task defined in Def.~\ref{node-level-motif-detection}.

\begin{lemma}
\label{prop:graph-classifier}
The node-level motif detection function $f_{node}: \mathcal{G}\rightarrow \{0,1\}^{|V|}$ contains sufficient information to solve the graph-level motif detection problem. Specifically, motif M exists in graph $G$ if and only if $\|\textbf{y}\|_0>0$, where $\textbf{y}=f_{node}(G)$ and $\| \cdot\|_0$ denotes the $\ell_0$ norm.
\end{lemma}



We then show (informally) below that, if there exists an ideal encoder $E_\text{node}: \{\textbf{f}_i^{(0)}\}_{i=1}^{|V|} \to \{\textbf{f}_i^{(t)}\}_{i=1}^{|V|}$ for node-level motif detection, it is not aligned with the frequency-dominated dynamics.

\begin{theorem}
	\label{thm:motif_detection_mixed_frequency}
The frequency-dominated regimes are suboptimal for node-level motif detection tasks.
\end{theorem}

The proofs for the above results are given in the Appendix~\ref{all_proofs}.
We show that effective node-level motif detection requires GNNs to be flexible to multiple frequency bands, which help us understand the impact of heterophily on graph-level tasks, and assist us to design new models for graph classification.

\paragraph{Implications for Heterophilic Graph Classification}
Our theoretical analysis reveals why heterophily impacts graph-level tasks differently than node-level tasks. In node classification, heterophily typically manifests as a spatial-consistency graph property that can be partially addressed through HFD dynamics. However, in motif-based graph classification, heterophily patterns are \textit{task-dependent} and spatially localized, creating three distinct challenges as mentioned in~\ref{sec:taxonomy}.

In a fine-grain clarification, we further divide the heterophily in motif-detection scenario into two genres: \textbf{Motif-agnostic heterophily} emerges when the background graph exhibits connectivity patterns independent of motif detection requirements, and \textbf{Motif-specific heterophily} arises because discriminative signals often require heterophilic patterns at motif boundaries, regardless of the background graph's structure. \textbf{Heterophily interference} occurs when mismatched patterns between motifs and backgrounds (\eg{} heterophilic motifs embedded in homophilic backgrounds) create conflicting optimization objectives for frequency-dominated approaches. This analysis directly motivates our experimental design in Section 4, where we systematically evaluate all four combinations of motif and background heterophily patterns to validate that mixed-frequency architectures outperform frequency-dominated approaches across diverse heterophily configurations.

\section{Experimental Validation}
In this section, we will verify our claims on both synthetic and real-world
datasets to support the theoretical analysis. For both synthetic and real-world experiments, we use three different GNN models, standard GCN
\cite{gilmer2017neural}, gradient flow GCN (GCN$_{gf}$) \cite{digiovanni2023understanding} and Adaptive Channel Mixing GNN (ACM-GNN) \cite{luan2022revisiting}, where standard GCN and GCN$_{gf}$ will provably lead to the frequency-dominated dynamic which enhances either low-frequency or high-frequency signals, while ACM-GNN are designed to adaptively combine multiple frequency filters. We make the parameters of these three models in the similar level to ensure them comparable. The train, validation and test sets are split by
80\%, 10\% and 10\% in both synthetic and real-world experiments.
The optimizer is Adam and the learning rate is set to 0.01 and 0.001 for synthetic and real-world experiments, respectively.

\begin{figure}[tb]
	\centering
	\includegraphics[width=\textwidth]{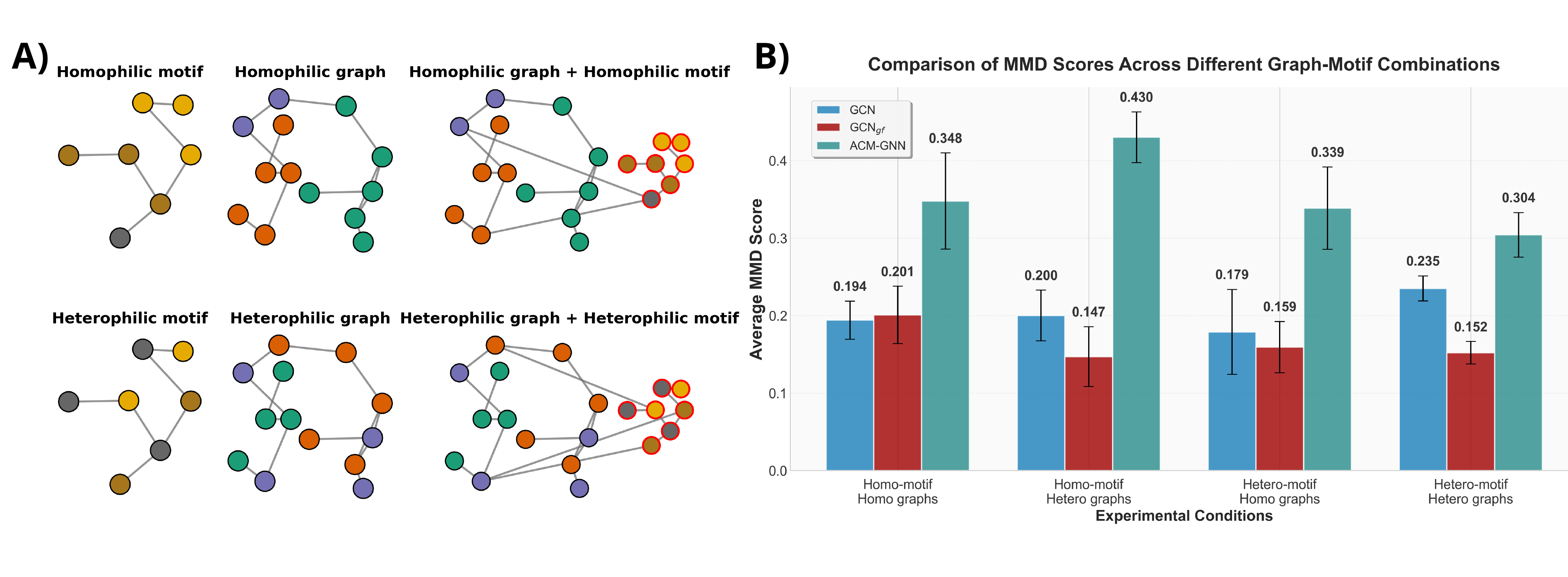}
    \vspace{-25pt}
	\caption{\textbf{(A)} The motifs and background graphs in two different settings (homophily and heterophily). Node colors represent different node features/labels. The example illustrates motifs and backbones with distinct feature distributions.
    \textbf{(B)} MMD scores of different conditions. The MMD scores are calculated on the test sets between graphs with and without motifs attached.}
	\label{fig:MMD_scores}
    \vspace{-14pt}
\end{figure}

\subsection{Synthetic Experiment}

We conduct experiments on four synthetic dataset variants, each representing a different combination of backbone and motif connectivity patterns: homophilic-homophilic, homophilic-heterophilic, heterophilic-homophilic, and heterophilic-heterophilic (see Fig.~\ref{fig:MMD_scores}(A) for demonstration and Appendix~\ref{synthetic_dataset_details} for details). For each dataset variant, we train all three GNN models and evaluate the best-performing model (selected via validation) on the corresponding test set.

We employ the empirical Maximum Mean Discrepancy (MMD) to quantify how effectively each GNN learns to distinguish between graph embeddings of different classes:
\begin{equation}
	\widehat{\text{MMD}}^2_\kappa(\{\mathbf{h}_i\}, \{\mathbf{g}_j\}) = \frac{1}{p^2}\sum_{i,i'=1}^{p} \kappa(\mathbf{h}_i, \mathbf{h}_{i'}) + \frac{1}{q^2}\sum_{j,j'=1}^{q} \kappa(\mathbf{g}_j, \mathbf{g}_{j'}) - \frac{2}{pq}\sum_{i=1}^{p}\sum_{j=1}^{q} \kappa(\mathbf{h}_i, \mathbf{g}_j)
\end{equation}
where $\{\mathbf{h}_i\}_{i=1}^{p}$ and $\{\mathbf{g}_j\}_{j=1}^{q} \in \mathbb{R}^d$ represent $p$ and $q$ final graph embeddings randomly sampled from graphs with and without motifs in the test set, respectively, and $\kappa$ denotes the RBF kernel function.

Figure~\ref{fig:MMD_scores}(B) presents the MMD scores across different experimental conditions. Higher MMD scores indicate superior discriminative capability between graphs containing motifs versus those without, while lower scores suggest diminished separation ability. Across all four scenarios, ACM-GNN consistently achieves higher MMD scores compared to the baseline models, while the two frequency-dominated approaches exhibit similar performance levels. These results validate our theoretical claims regarding the limitations of frequency-dominated GNNs for motif detection tasks.

\subsection{Real-world Experiment}


\begin{table}[t]
\centering
\caption{Results on the pK$_a$ dataset and Dirichlet Energy analysis.}
\begin{tabular}{cccccc}
\toprule
\multicolumn{2}{c}{Performance} & \multicolumn{4}{c}{Shrink Ratio of Normalized Dirichlet Energy} \\
\cmidrule(r){1-2} \cmidrule(l){3-6}
Model & MSE $\downarrow$ & Boundary $\uparrow$ & R$_{2}$-NH & R-CH=O & R-C(=NH)NH2 \\
\midrule
GCN       & $3.08 \pm 0.31$ & $0.14 \pm 0.01$ & $0.13 \pm 0.01$ & $0.17 \pm 0.02$ & $0.16 \pm 0.03$ \\
GCN$_{gf}$ & $3.24 \pm 0.40$ & $0.13 \pm 0.01$ & $0.19 \pm 0.01$ & $0.12 \pm 0.01$ & $0.11 \pm 0.02$ \\
ACM-GNN   & $\textbf{2.32} \pm \textbf{0.38}$ & $1.77 \pm 0.17$ & $0.24 \pm 0.08$ & $0.14 \pm 0.01$ & $0.18 \pm 0.04$ \\
\bottomrule
\end{tabular}
\label{tab:real-world-results}
\end{table}

To practically verify our claim on real-world graph-level tasks, we evaluated the baseline models on a newly collected dataset (see Appendix \ref{pka_dataset_details} for details). The dataset comprises 6,714 molecules with their corresponding pK$_a$ (acid-base dissociation constant) values. The pK$_a$ value quantifies the acidity or basicity of a molecule and is strongly influenced by specific functional groups, which correspond to distinctive motifs in the molecular graph structure. This constitutes a graph-level regression task where molecules serve as input and the objective is to predict the pK$_a$ value of them.

Table \ref{tab:real-world-results} reports the experimental results. The mean squared error (MSE) is evaluated on the test set and averaged over five independent runs. Consistent with the synthetic data experiments, ACM-GNN achieves the best performance among the three GNN models. We further compute the shrink ratio of the normalized Dirichlet energy, 
$
\frac{\mathcal{E}^{Dir}(\mathbf{F}(t))/\|\mathbf{F}(t)\|^2}{\mathcal{E}^{Dir}(\mathbf{F}(0))/\|\mathbf{F}(0)\|^2}
$
where \(t\) denotes the evolution time of the dynamical system (\ie{} the number of layers). Since Dirichlet energy reflects differences across edges, we focus on two categories: boundary edges (those connecting motifs to the backbone graph) and intra-motif edges (within the three functional groups). The results show that GCN and GCN\(_{gf}\) yield low shrink ratios for both boundary and intra-motif edges, indicating global over-smoothing consistent with LFD dynamics. In contrast, ACM-GNN produces high shrink ratios on boundary edges and low shrink ratios within motifs, effectively sharpening boundaries while smoothing internal embeddings. This behavior explains the observed performance gap across models and supports our hypothesis on heterophily patterns in motif detection (Sec.~\ref{heteorphily_patterns_in_motif_detection}).  


\section{Conclusion and Future Work}
Our theoretical and empirical analysis demonstrates that effective motif detection demands a spatially adaptive dynamic, rejecting monolithic low- or high-frequency dominated (LFD/HFD) regimes. Successful models must resolve a fundamental tension: performing intra-motif smoothing to unify constituent nodes while simultaneously sharpening boundaries to distinguish the motif from the wider graph structure. Our energy-based framework formalizes why globally frequency-dominated dynamics are ill-suited for this, revealing that their asymptotic convergence actively destroys the spectral signatures required to identify local patterns.

This energy-based perspective opens several avenues for future research. First, a deeper characterization of motif-specific dynamics promises to guide the principled design of specialized GNN architectures that excel at local structural pattern recognition. Second, extending our dynamic-based taxonomy to a broader class of graph-level tasks---and to other architectures like Graph Transformers---could provide a unified theory explaining performance variations across different models and problem settings. Such an analysis would clarify the nuanced role of heterophily at both the node and graph levels, paving the way for more robust and versatile graph learning models.
\newpage

\section*{Acknowledgement}
This study was funded by the European Union (DTRIP4H, No. 101188432) and iCANDOC Precision Cancer Medicine (PCM) pilot program from the Research Council of Finland.

\bibliographystyle{plain}
\bibliography{references}

\newpage
\appendix
\section{Real-world Examples for Graph-level Task Taxonomy}
\label{ref:real-world-examples}
\subsection{Examples for Aggregated Node Feature Based Labeling }

\begin{itemize}
    \item \textbf{Molecular property prediction}: Classifying drug molecules based on average atomic properties (\textit{\eg{}} if the average electronegativity of atoms exceeds a threshold, classify as ``polar'' vs ``non-polar'')~\cite{mao_study_2025, tantardini_thermochemical_2021}
    
    \item \textbf{Social network analysis}: Categorizing online communities based on aggregated user demographics (\textit{\eg{}} if $>70\%$ of users are in a certain age group, label the network as ``young adult community'')~\cite{traud2012social, perozzi2014deepwalk}
    
    \item \textbf{Brain network analysis}: Classifying brain connectivity networks based on average activation levels across brain regions (\textit{\eg{}} networks with high average activity labeled as ``hyperactive state'')~\cite{bullmore_complex_2009, rahaman_deep_2024}
    
    \item \textbf{Protein interaction networks}: Classifying protein complexes based on the proportion of proteins belonging to specific functional categories~\cite{kovacs_network-based_2019,hua2024mudiff}
\end{itemize}

\subsection{Substructure labeling examples}

\begin{itemize}
    \item \textbf{Drug discovery}: Classifying molecules based on the presence of specific pharmacophores or toxic substructures (\textit{\eg{}} presence of benzene rings, specific functional groups)~\cite{bongini2021molecular, gaudelet2021utilizing}
    
    \item \textbf{Social network analysis}: Detecting communities based on local clustering patterns -- networks with many tightly-knit triangular relationships vs. those with more heterophilic connections~\cite{tang2009social, fortunato_community_2010}
    
    \item \textbf{Transportation networks}: Classifying road networks based on the presence of specific traffic patterns like roundabouts, highway interchanges, or bottleneck structures~\cite{boyles_transportation_2025}
    
    \item \textbf{Chemical reaction networks}: Categorizing reaction pathways based on the presence of specific reaction motifs or catalytic cycles~\cite{hua2024reactzyme, wen_chemical_2023}
\end{itemize}

\subsection{Global labeling examples}

\begin{itemize}
    \item \textbf{Molecular classification}: Distinguishing between different molecular families based on overall structural properties like molecular diameter, overall connectivity, or graph density~\cite{wu2018moleculenetbenchmarkmolecularmachine}
    
    \item \textbf{Social network analysis}: Classifying entire social networks based on global properties like average path length (small-world vs. random networks) or overall network density~\cite{hu2020open, wang_microsoft_2020}
    
    \item \textbf{Infrastructure networks}: Classifying power grids or communication networks based on their overall robustness, measured by global connectivity metrics~\cite{gong2023beyond}
\end{itemize}

\section{The Synthetic Dataset}
\label{synthetic_dataset_details}
Our synthetic dataset consists of graphs composed of a larger background graph (backbone) with a smaller substructure (motif) embedded within it. We generate backbone graphs and motifs independently, with each type designed to exhibit either homophilic or heterophilic node connectivity patterns.

\textbf{Dataset Composition.} We generate 1,000 backbone graphs for each connectivity type (homophilic and heterophilic) and create 5 distinct motif types, also categorized by connectivity pattern. Each backbone graph is paired with one motif to form a complete graph, resulting in four possible combinations: (homophilic backbone, homophilic motif), (homophilic backbone, heterophilic motif), (heterophilic backbone, homophilic motif), and (heterophilic backbone, heterophilic motif). With 1,000 backbone graphs and 5 motif variants per combination, each of the four combinations contains 5,000 graphs, yielding a total dataset of 20,000 graphs.

\textbf{Graph Generation Process.} The synthesis procedure consists of two stages: (1) structural generation of unlabeled graph skeletons, and (2) node feature assignment. The skeleton is first constructed by adding extra edges to a random tree, where the number of added edges is preset to half of the maximum possible edges for the given number of nodes. For homophilic graphs, node labels are assigned using the Clauset–Newman–Moore greedy modularity maximization algorithm~\cite{PhysRevE.70.066111}, which encourages intra-community similarity. For heterophilic graphs, node labels are instead assigned at random to promote dissimilarity among connected nodes. The initial node features $\mathbf{f}_0$ are generated using a fixed embedding layer in PyTorch with small perturbations (gaussian noise with $\mu=0$ and $\sigma=0.05$), ensuring that nodes sharing the same label have similar features.

\textbf{Graph Constraints.} Backbone graphs contain 20--50 nodes, while motifs contain 5--7 nodes. And the edges are randomly sampled from minimum ($|V|-1$) to maximum ($\frac{|V|\times(|V|-1)}{4}$). The maximum number of edges is set to ensure reasonable connectivity without creating overly dense graph.

\section{The pK$_a$ Dataset}
\label{pka_dataset_details}
\begin{figure}
    \centering
    \includegraphics[width=\linewidth]{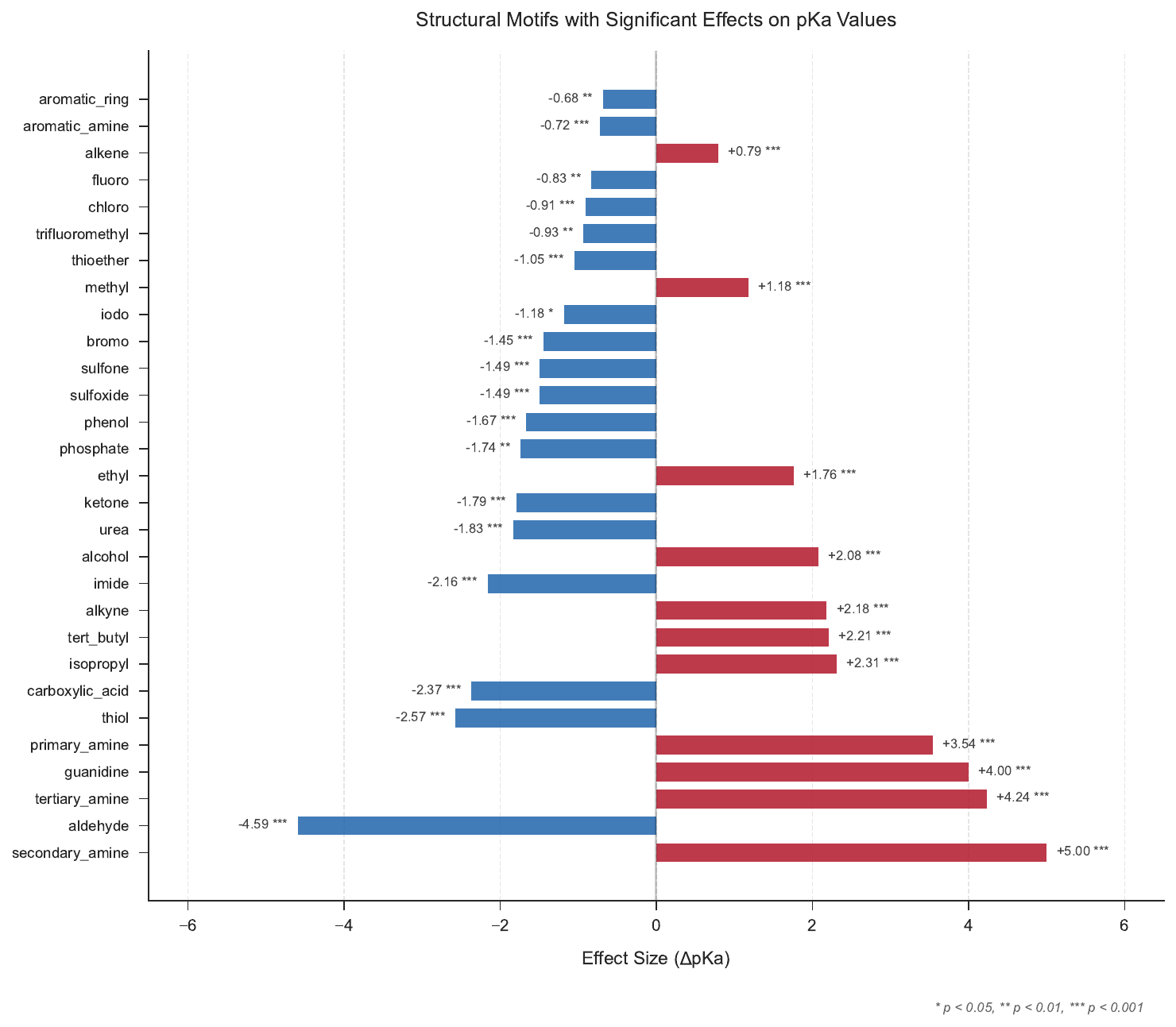}
    \caption{The influence of the functional groups on the pK$_{a}$ values. The y-axis refers to different functional groups, and the x-axis refers to the change of the pK$_{a}$ value if this specific functional group appears in the molecule.}
    \label{fig:motif_pka_correlation}
\end{figure}
The dataset is collected from the IUPAC Dissociation Constants GitHub repository \footnote{\url{https://github.com/IUPAC/Dissociation-Constants}}, which serves as a publicly accessible, continuously updated resource containing high-confidence pK$_a$ data. This dataset has been meticulously digitized and curated from authoritative reference works published by the International Union of Pure and Applied Chemistry (IUPAC), ensuring data quality and reliability for computational chemistry applications.

From the total 24,290 records available in the repository, we select 6,714 unique molecules with their corresponding pK$_a$H$_1$ values, which represent the equilibrium constant for the loss of the first proton from each molecule. This subset focuses on molecules with well-defined ionization properties, making it suitable for studying the relationship between molecular structure and chemical reactivity. The pK$_a$ values in our dataset span a wide range of chemical environments, encompassing various functional groups and molecular frameworks encountered in pharmaceutical and chemical research. In Fig.~\ref{fig:motif_pka_correlation} we list the influence of functional groups on the molecule's pK$_a$.

For our task, we convert each molecule into a graph representation where atoms serve as nodes and chemical bonds as edges. Node features include atomic properties, \eg{} atomic number, formal charge, and hybridization state, while edge features capture bond types and stereochemistry information.

Table~\ref{tab:pka_dataset} presents statistical properties of the pK$_a$ values in our curated dataset, including the distribution range, mean, median, and standard deviation, providing insight into the chemical diversity and complexity of the molecular structures under investigation.
\begin{table}[tb]
    \centering
    \caption{Statistical properties of the pK$_a$ value in the dataset.}
    \begin{tabular}{lccc}
    \toprule
      \textbf{Average} &  \textbf{Standard Variance} & \textbf{Maximum} & \textbf{Minimum} \\
    \midrule
      4.063  &  4.259 & 14.110 & -17.632\\
    \bottomrule
    \end{tabular}
    \label{tab:pka_dataset}
\end{table}

\section{Equation Derivation}
\label{equal_derivation}
\subsection{Eigenvectors Decomposition}
\label{eigen_decomposition}
Since $\textbf{W} \in \mathbb{R}^{n\times n}$ is symmetric, we write it as $\textbf{W}=\textbf{Q}\mathbf{\Lambda}\textbf{Q}^\top$, where $\mathbf{\Lambda}$ is a digonal matrix with all the eigenvalues $\boldsymbol{\mu} \in \{\boldsymbol{\mu}_0, \boldsymbol{\mu}_1, \dots, \boldsymbol{\mu}_{n-1}\}$. Suppose $\boldsymbol{\mu}_k$ is the smallest positive eigenvalue, and $\boldsymbol{\mu}_0 \le \boldsymbol{\mu}_1 \le \dots \le \boldsymbol{\mu}_k \le \dots \le \boldsymbol{\mu}_{n-1}$, we have:
\begin{equation*}
\begin{aligned}
\mathbf{\Lambda} & =
\begin{bmatrix}
        \boldsymbol{\mu}_0 & 0 & \dots & 0 \\
        0 & \boldsymbol{\mu}_1 & \dots & 0 \\
        \vdots & \vdots & \ddots & 0 \\
        0 & 0 & \dots & \boldsymbol{\mu}_{n-1}
\end{bmatrix}
\\ & =
\begin{bmatrix}
        \boldsymbol{\mu}_0 & 0 & \dots & \dots & \dots &0 \\
        0 & \boldsymbol{\mu}_1 & \dots & \dots & \dots & 0 \\
        \vdots & \vdots & \ddots & \dots & \dots &0 \\
        0 & 0 & \dots & \boldsymbol{\mu}_{k-1} &\dots & 0 \\
        \vdots & \vdots & \dots  &\dots &\ddots & 0 \\
        0 & 0 & \dots &\dots &\dots & 0
\end{bmatrix}
+
\begin{bmatrix}
        0 & 0 & \dots & \dots & \dots &0 \\
        0 & 0 & \dots & \dots & \dots & 0 \\
        \vdots & \vdots & \ddots & \dots & \dots &0 \\
        0 & 0 & \dots & \boldsymbol{\mu}_{k} &\dots & 0 \\
        \vdots & \vdots & \dots  &\dots &\ddots & 0 \\
        0 & 0 & \dots &\dots &\dots & \boldsymbol{\mu}_{n-1}
\end{bmatrix}
\\ & =
\mathbf{\Lambda}_{+} - \mathbf{\Lambda}_{-}
\end{aligned}
\end{equation*}
where $\mathbf{\Lambda}_{+}$ is a diagonal matrix with all the positive eigenvalues (from $\boldsymbol{\mu}_k$ to $\boldsymbol{\mu}_{n-1}$) and $\mathbf{\Lambda}_{-}$ is a diagonal matrix with absolute values of all the negative eigenvalues (from $\boldsymbol{\mu}_0$ to $\boldsymbol{\mu}_{k-1}$).
We note $\sqrt{\mathbf{\Lambda}_{+}}$ and $\sqrt{\mathbf{\Lambda}_{-}}$ as the element-wise square root of the two matrices $\mathbf{\Lambda}_{+}$ and $\mathbf{\Lambda}_{-}$ respectively.
Furthermore, we define $\mathbf{\Theta}_{+}=\sqrt{\mathbf{\Lambda}_{+}}\textbf{Q}^{\top}$ and $\mathbf{\Theta}_{-}=\sqrt{\mathbf{\Lambda}_{-}}\textbf{Q}^{\top}$.
Then we can rewrite the weight matrix $\textbf{W}$ as:
\begin{equation*}
    \begin{aligned}
        \textbf{W} & = \textbf{Q}(\mathbf{\Lambda}_{+} - \mathbf{\Lambda}_{-})\textbf{Q}^{\top}\\
         & = \textbf{Q}\mathbf{\Lambda}_{+}\textbf{Q}^{\top} - \textbf{Q}\mathbf{\Lambda}_{-}\textbf{Q}^{\top} \\
         & = \textbf{Q}(\sqrt{\mathbf{\Lambda}_{+}})(\sqrt{\mathbf{\Lambda}_{+}})\textbf{Q}^{\top} - \textbf{Q}(\sqrt{\mathbf{\Lambda}_{-}})(\sqrt{\mathbf{\Lambda}_{-}})\textbf{Q}^{\top} \\
         & = (\textbf{Q}\sqrt{\mathbf{\Lambda}_{+}})(\sqrt{\mathbf{\Lambda}_{+}}\textbf{Q}^{\top}) - (\textbf{Q}\sqrt{\mathbf{\Lambda}_{-}})(\sqrt{\mathbf{\Lambda}_{-}}\textbf{Q}^{\top})\\
         & = \mathbf{\Theta}_{+}^{\top}\mathbf{\Theta}_{+}-\mathbf{\Theta}_{-}^{\top}\mathbf{\Theta}_{-}
    \end{aligned}
\end{equation*}

\section{Proofs}
\label{all_proofs}
\subsection{Proof of Proposition \ref{equivalence_between_tasks}}
\begin{proof}
    \label{proof:equivalence}
    We prove both directions of the equivalence.
    \paragraph{(i) $\Rightarrow$ (ii)}
    	Assume graph $G$ contains a subgraph isomorphic to motif $M$. That is, there exists a subgraph $H = (V_H, E_H)$ where $V_H \subseteq V$, $E_H \subseteq E$, and $H \cong M$.
	
	Since $H$ is isomorphic to $M$, by definition, every node $i \in V_H$ is part of a subgraph in $G$ that is isomorphic to $M$ (namely, $H$ itself). Therefore, by Definition~\ref{node-level-motif-detection}, for every node $i \in V_H$, we have $y_i = 1$.
	
	Since $V_H \neq \emptyset$ (as $M$ is a non-empty motif), there exists at least one node $i \in V$ such that $y_i = 1$.

    \paragraph{(ii) $\Rightarrow$ (i)}
    	Assume there exists at least one node $i \in V$ such that $y_i = 1$. 
	
	By Definition~\ref{node-level-motif-detection}, $y_i = 1$ means that node $i$ is part of some subgraph in $G$ that is isomorphic to $M$. Let $H$ denote this subgraph. Then $H \subseteq G$ and $H \cong M$.
	
	Therefore, graph $G$ contains a subgraph isomorphic to motif $M$.

    \paragraph{Conclusion}
    Thus, the equivalence (i) $\Leftrightarrow$ (ii) is established. This proves that the objective of distinguishing nodes in a substructure (node-level motif detection) is equivalent to detecting the existence of the substructure (graph-level motif detection) in the sense that:
	$$\exists \text{ subgraph } H \subseteq G : H \cong M \quad \Leftrightarrow \quad \exists i \in V : y_i = 1$$
\end{proof}

\subsection{Proof of Lemma \ref{prop:graph-classifier}}
\begin{proof}
\label{proof:graph_classifier}
To prove this, we must show two things: (1) a perfect node-level classifier contains sufficient information to construct a perfect graph-level classifier, and (2) a perfect graph-level classifier does not contain sufficient information to construct a perfect node-level classifier.

Let $G=(V, E)$ be a graph and $M$ be a target motif. 
The node-level task requires learning a function $f_{node}: \mathcal{G} \to \{0, 1\}^{|V|}$ that predicts the label vector $\mathbf{y} = (y_1, \dots, y_{|V|})$, where $y_i=1$ if node $i$ is part of a subgraph isomorphic to $M$. The graph-level task requires learning a function $f_G: G \to \{0, 1\}$ that predicts a single label $y_G=1$ if $G$ contains any subgraph isomorphic to $M$.

\paragraph{A node-level solution implies a graph-level solution.}

Given a perfect node-level classifier $f_{node}$, we can construct a perfect graph-level classifier, $f_G^*$, using a simple transformation $T: \{0, 1\}^{|V|} \to \{0, 1\}$:
\[
    f_G^*(G) = T(f_{node}(G)) = \max(f_{node}(G)).
\]
By definition, a graph $G$ contains a motif $M$ if and only if there is at least one node $i \in V$ that is part of a subgraph isomorphic to $M$. This is equivalent to the condition that at least one entry in the true node-level label vector $\mathbf{y}$ is 1. The function $f_G^*$ correctly implements this, as $\max(\mathbf{y}) = 1$ if and only if $\sum y_i > 0$. Thus, the information provided by $f_{node}$ is sufficient to solve the graph-level task.

\paragraph{A graph-level solution does not imply a node-level solution.}

We prove this by counterexample. Let the motif $M$ be a triangle ($K_3$) and consider two graphs, $G_1$ and $G_2$, on the vertex set $V=\{v_1, v_2, v_3, v_4\}$.
\begin{itemize}
    \item Let $G_1$ be a graph consisting of a single triangle on nodes $\{v_1, v_2, v_3\}$ and an isolated node $v_4$.
    \item Let $G_2$ be a complete graph ($K_4$) on all four nodes.
\end{itemize}
A perfect graph-level classifier $f_G$ will produce the same output for both, as both graphs contain at least one triangle:
\[
    f_G(G_1) = 1 \quad \text{and} \quad f_G(G_2) = 1.
\]
However, the ground-truth node-level label vectors are different:
\begin{itemize}
    \item For $G_1$, the node-level vector is $\mathbf{y}_1 = (1, 1, 1, 0)$.
    \item For $G_2$, the node-level vector is $\mathbf{y}_2 = (1, 1, 1, 1)$, since all nodes in a $K_4$ are part of at least one triangle.
\end{itemize}
Any transformation attempting to construct a node-level classifier from the graph-level output would need to map the input value `1` to two different outputs, $\mathbf{y}_1$ and $\mathbf{y}_2$. This is impossible for a function. Therefore, the information provided by a graph-level classifier is fundamentally insufficient to solve the node-level task, as it cannot distinguish between different node configurations that satisfy the same graph-level property. Since a node-level solution implies a graph-level solution but the reverse is not true, the node-level task is strictly more general.
\end{proof}

\subsection{Proof of Theorem \ref{thm:motif_detection_mixed_frequency}}

\begin{figure}[tb]
    \centering
    \includegraphics[width=1.0\linewidth]{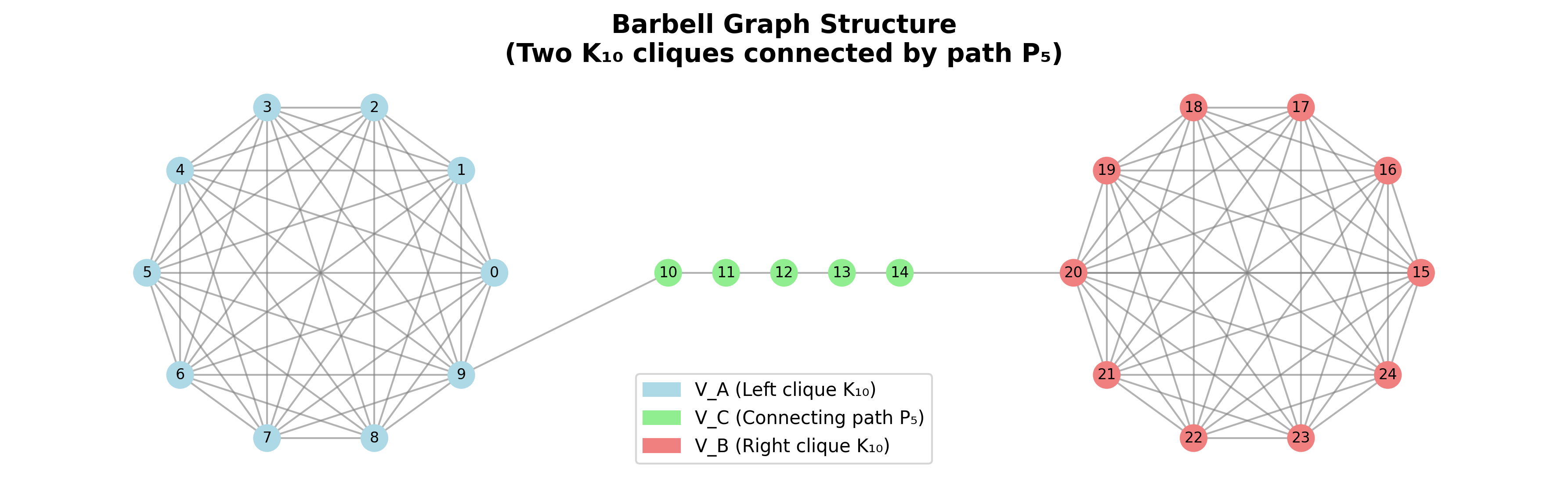}
    \caption{A barbell-shape graph consists of two complete graph $K_{10}$ and a connection path line graph $P_5$.}
    \label{fig:barbell_graph}
\end{figure}

\begin{proof}
	\label{proof:motif_detection_mixed_frequency}
	Let $\mathcal{L}_{motif}$ denote the motif detection loss function, and let $\mathbf{f}^*_M$ denote the optimal node feature representation for motif detection.
	We analyze how the energy functional $\mathcal{E}_{\mathbf{\theta}}(\mathbf{F})$ from Eq.~\ref{general_energy} conflicts with motif detection objectives under LFD and HFD regimes.
	
	\paragraph{LFD Suboptimality} 
	According to the framework in Section 2, LFD dynamics minimize the normalized Dirichlet energy $\mathcal{E}^{Dir}(\mathbf{F}(t))/\|\mathbf{F}(t)\|^2$, driving the system toward the global minimum where $\mathbf{F}(t) \to c\boldsymbol{\phi}_0 \mathbf{1}^T$ ($c$ is the proportional coefficient).
	
	In the gradient flow formulation $\dot{\mathbf{F}}(t) = -\nabla \mathcal{E}_{\mathbf{\theta}}(\mathbf{F}(t))$, the pairwise interaction term $-\sum_{i,j} A_{ij} \langle \mathbf{f}_i, \mathbf{W} \mathbf{f}_j \rangle$ encourages neighboring nodes to have aligned representations (when $\mathbf{W}$ has positive eigenvalues). This global alignment objective is fundamentally at odds with motif detection.
	
	For motif detection, the optimal loss $\mathcal{L}_{motif}$ requires representations to distinguish between motif-participating nodes $V_M$ and non-participating nodes $V \setminus V_M$. However, LFD dynamics optimize for global consensus, producing identical representations $\mathbf{f}_i \approx \mathbf{f}_j$ for all $i,j \in V$ (\eg{} the node on the boundary like node 9 and node 10 in the Fig.~\ref{fig:barbell_graph}). This makes the model to have blurred edges between motifs and the background graph and eliminates the discriminative information necessary for motif detection, resulting in $\mathcal{L}_{motif}(c\boldsymbol{\phi}_0 \mathbf{1}^T) > \mathcal{L}_{motif}(\mathbf{f}^*_M)$.
	
	\paragraph{HFD Suboptimality}
	From the theorem in Section 2, HFD dynamics occur when $|\boldsymbol{\mu}_0|(\boldsymbol{\lambda}_{n-1}-1)>\boldsymbol{\mu}_{d-1}$, leading to maximization of the normalized Dirichlet energy and convergence to $\mathbf{F}(t) \to c\boldsymbol{\phi}_{n-1} \mathbf{1}^T$.
	
	In this regime, the energy functional's pairwise term $-\sum_{i,j} A_{ij} \langle \mathbf{f}_i, \mathbf{W} \mathbf{f}_j \rangle$ drives neighboring nodes to have maximally different representations, as the system seeks to maximize $\sum_{(i,j) \in E} \|\mathbf{f}_i - \mathbf{f}_j\|^2$.
	
	This creates a fundamental conflict with motif detection objectives. Consider any connected motif $M$ - nodes within the same motif instance are connected by edges and should receive similar labels (both should be classified as motif participants). However, HFD dynamics force these neighboring nodes to have anti-aligned representations: $\mathbf{f}_i \approx -\mathbf{f}_j$ for $(i,j) \in E$, \eg{} nodes in the complete graph $K_{10}$ on the sides in Fig.~\ref{fig:barbell_graph}.
	
	The energy minimization process actively works against the motif detection objective, making it impossible for any linear classifier to consistently label connected nodes within the same motif instance. Thus $\mathcal{L}_{motif}(c\boldsymbol{\phi}_{n-1} \mathbf{1}^T) > \mathcal{L}_{motif}(\mathbf{f}^*_M)$.
	
	\paragraph{Mixed-Frequency Requirement}
	The energy framework reveals why motif detection is incompatible with frequency-dominated regimes. The gradient flow $\dot{\mathbf{F}}(t) = -\nabla \mathcal{E}_{\mathbf{\theta}}(\mathbf{F}(t))$ drives the system toward eigenvector alignment, but motif detection requires a different objective function altogether.
	
	Optimal motif detection requires minimizing $\mathcal{L}_{motif}$, which demands:
	\begin{itemize}
		\item \textbf{Inner smoothing}: Similar representations within motif instances, \ie{} local low-frequency behavior
		\item \textbf{Boundary sharpening}: Sharp boundaries between motif and non-motif regions, \ie{} local high-frequency behavior
		\item \textbf{Scale-appropriate sensitivity}: Frequency components matching the motif's characteristic size
	\end{itemize}
	
	This creates a mixed-frequency optimization problem that cannot be solved by the global energy minimization of $\mathcal{E}_{\mathbf{\theta}}$. The optimal solution $\mathbf{f}^*_M$ requires spatially-varying frequency content: $\mathbf{f}^*_M = \sum_{k} \alpha_k \boldsymbol{\phi}_k \mathbf{v}_k$ where the coefficients $\alpha_k$ depend on the local structural context around each motif.
	
	Since the energy functional in Eq.~\ref{general_energy} enforces global spectral alignment (either LFD or HFD), it cannot accommodate the spatially-heterogeneous frequency requirements of motif detection. Therefore, effective motif detection requires architectures that can escape the LFD/HFD dichotomy imposed by the standard energy framework.
\end{proof}
\end{document}